\documentclass[sigconf]{acmart}

\calclayout

\renewcommand\footnotetextcopyrightpermission[1]{} 
\AtBeginDocument{%
  \providecommand\BibTeX{{%
    \normalfont B\kern-0.5em{\scshape i\kern-0.25em b}\kern-0.8em\TeX}}}

\usepackage{hyperref}
\usepackage{graphicx}
\usepackage{caption}
\usepackage{subcaption}
\usepackage{amsfonts}
\usepackage{float}
\usepackage{dblfloatfix}
\usepackage{enumitem}

\usepackage{graphicx}
\usepackage{dcolumn}
\usepackage{bm}

\graphicspath{ {./figures/} }

\copyrightyear{2022}
\acmYear{2022}
\setcopyright{rightsretained}
\acmConference[CHASE' 22]{ACM/IEEE International Conference on
Connected Health: Applications, Systems and Engineering
Technologies}{November 17--19, 2022}{Washington, DC, USA}
\acmBooktitle{ACM/IEEE International Conference on Connected Health:
Applications, Systems and Engineering Technologies (CHASE' 22),
November 17--19, 2022, Washington, DC, USA}
\acmDOI{10.1145/3551455.3564708}
\acmISBN{978-1-4503-9476-5/22/11}

\begin{document}
\title{Shoupa: An AI System for Early Diagnosis of Parkinson’s Disease}




\author{
Jingwei Li$^1$, 
Ruitian Wu$^1$,
Tzu-liang Huang$^1$, 
Zian Pan$^1$,
Ming-Chun Huang$^{1, 2}$}


\thanks{
\begin{enumerate}[leftmargin=*,topsep=0pt, label={\arabic*}.]
\item \textit{ Jingwei Li, Ruitian Wu, Tzu-liang Huang, Zian Pan, and Mingchun Huang are with Duke Kunshan University, China. (Equal Contribution with alphabetical order)} \textit{E-mail: \{ jl957, rw219, th270, zp45, mh596\}@duke.edu.}
\item \textit{Ming-Chun Huang is also with Suzhou Huanmu Intelligence Technology Co., Ltd, China}
\end{enumerate}

Permission to make digital or hard copies of part or all of this work for personal or classroom use is granted without fee provided that copies are not made or distributed for profit or commercial advantage and that copies bear this notice and the full citation on the first page. Copyrights for third-party components of this work must be honored. For all other uses, contact the owner/author(s).

\textit{CHASE' 22, November 17-19, 2022, Washington, DC, USA }

\textcopyright 2022 Copyright held by the owner/author(s).

ACM ISBN 978-1-4503-9476-5/22/11. \url{https://doi.org/10.1145/3551455.3564708}}



\begin{abstract}
    Parkinson’s Disease (PD) is a progressive nervous system disorder that has affected more than 5.8 million people, especially the elderly. Due to the complexity of its symptoms and its similarity to other neurological disorders, early detection requires neurologists or PD specialists to be involved, which is not accessible to most old people. Therefore, we integrate smart mobile devices with AI technologies. In this paper, we introduce the framework of our developed PD early detection system which combines different tasks evaluating both motor and non-motor symptoms. With the developed model, we help users detect PD punctually in non-clinical settings and figure out their most severe symptoms. The results are expected to be further used for PD rehabilitation guidance and detection of other neurological disorders.
\end{abstract}

\maketitle
\pagestyle{plain}

\section{Introduction}
\
\indent Parkinson's disease (PD) is a progressive nervous system disorder associated with a decline in motor and non-motor physical function. In 2019, PD resulted in 5.8 million disability-adjusted life years and is estimated to influence 14.2 million people by 2040 \cite{dorsey2018parkinson}. Although PD cannot be cured entirely, early detection and treatment can be beneficial for people with PD. Detection in the early stages of PD not only helps relieve symptoms \cite{jankovic2008parkinson} but also prolongs life expectancy \cite{pinto2004treatments}.

However, the early detection of PD is often difficult since PD's early symptoms are subtle and it's similar to other neurological disorders of the brain \cite{postuma2015mds}. Although a traditional solution of getting a complete evaluation with the help of a neurologist or PD specialist at an early stage is helpful, more than 40\% of individuals older than 65 have fewer opportunities to meet a PD specialist, even in wealthy countries  \cite{dorsey2018parkinson}. The lack of opportunities may result from the strict testing environment and the time-consuming nature of the diagnostic procedures to screen out other possible diseases.

With the spread of smart mobile devices, early PD diagnosis can be conducted more easily and more accessible to patients. There are mainly three advantages of mobile detection. First, PD testing will no longer be influenced by time, space, and facility. With the help of mobile devices, the test can be conducted in a non-clinical setting. Second, mobile PD detection can provide good accuracy with the assistance of computers together with some existing techniques. Third, the use of mobile health allows a continuous collection and monitoring of patient data. As a chronic disease, constantly collected data would help trace the effectiveness of follow-up medication and physical rehabilitation for various symptoms.

\begin{figure*}
    \centering  
    \includegraphics[height=3.9cm,width=13cm]{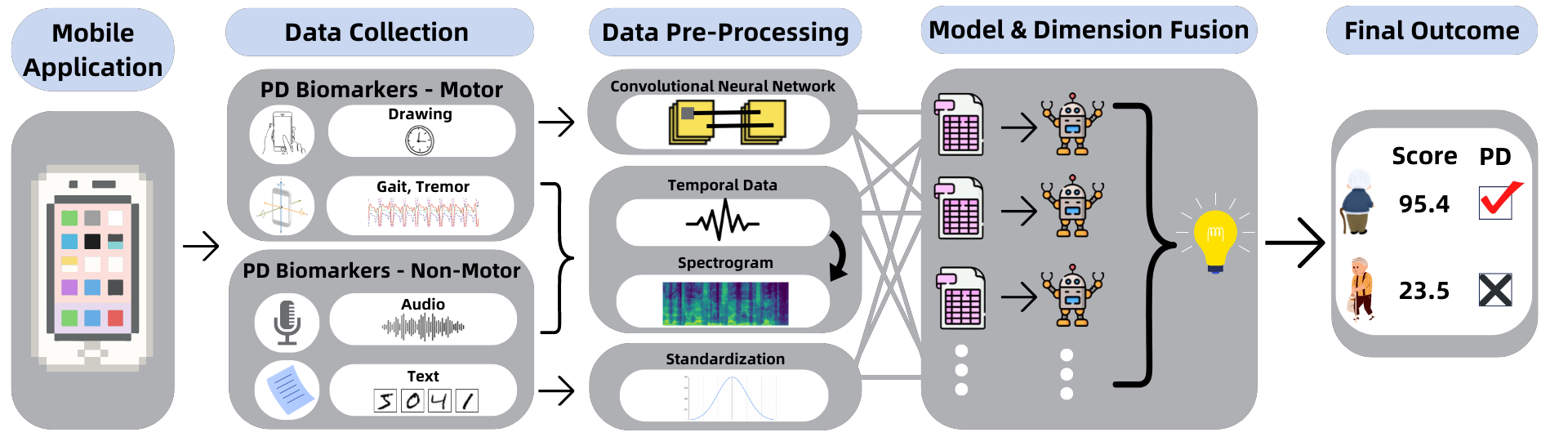}
    \caption{Shoupa System's Overall Workflow}
    \label{workflow_rhythm}
\end{figure*}

\section{System Design}

\
\indent In response to the above-mentioned inconveniences of the current Parkinson's disease early diagnosis, we develop a mobile Parkinson's diagnosis software enabling users to know whether they are at risk of having PD even at home. 

Shoupa is developed based on the Flutter framework and can be run on Android 5.0+ or IOS 9.0+ devices. Its Back-end is based on an Amazon AWS server. The system supports both motor and non-motor tests. In motor tests, we focus on the users' finger movement and body balance, in which finger movement is examined by rhythm and drawing tests, while body balance is examined by gait and tremor tests. Meanwhile, in non-motor tests, we measure users' memory and vocal ability and also collect user's general information through three surveys: MDS-UPDRS, MMSE, and demographic surveys. We designed visual memory and auditory memory tests to check users' memory and used pronunciation and reading tests for vocal ability examination. 

Since we have multiple categories of tests, the modality of data we receive also varies greatly, including gyroscope data (gait and tremor tests), audio data (pronunciation and reading tests), image data (drawing test), and numerical/categorical data (surveys). We will process data of the same modality in the same way to convert all the attributes into standardized 1-D vectors.

After pre-processing the data, we will fuse attributes describing the same body function into a scorer.
Based on the scorer, the system will give an evaluation result to that certain function (For example, finger movement only). Since our system tests multiple functions of the body, we will finally have several scorers. Finally, all the scorers will be ensembled with certain weights to give a comprehensive diagnosis of PD.

Figure \ref{workflow_rhythm} shows the overall workflow of our system.


\section{Data Collection and Analysis}
\
\indent In this section, methods that we used to collect and analyze data is introduced in greater detail.

To collect data on PD patients in China, We collaborate with Guangzhou First People's Hospital. Patients conduct our tests under the guidance of medical professionals so that the data is guaranteed to be real and valid. All the data being collected is desensitized and is only used for research purposes.

With the collected data, we build a detection model which could return how likely the users will be affected by PD and what their most severe symptoms are. The essence of the model is based on multi-modal and multi-category data:
\begin{enumerate}[topsep=0pt]
    \item Multi-modality: For all collected attributes, there could be data of different modalities, including numerical/ categorical data (surveys), image data (drawing test), gyroscope data (gait and tremor tests), and audio data (pronunciation and reading tests). 
    \begin{enumerate}
        \item For the basic numerical/categorical data, they are supposed to be standardized to mitigate the impacts of the magnitude of data.
        \item For the image data, we employ basic CNN with Residual Layers\cite{7780459} appended to transform into a 1-D vector.
        \item For the gyroscope data and audio data, since they are both temporal-based, we process them in the same way. We transform it firstly into 2D time-frequency spectrogram representation and then process them similarly to how we process 2-D data.
    \end{enumerate}
    
    \item Multi-dimension: We have data describing different functions, including speech, memory, body balance, etc. We directly fuse the multi-modal variables describing the same function as a whole dimension. The diverse dimensions of our model ensure that our model gives a comprehensive evaluation to the users. For each dimension, a scorer is trained based on Neural Network or Support Vector Machine. We expect to implement the multi-dimensional structure with XGBoost\cite{XGboost} to optimize weights for each scorer and employ it to reflect how each symptom contributes to the final diagnosis. 
\end{enumerate}

By integrating all different dimensions and modalities, we expect to give a final scorer giving a reliable diagnosis of the PD. Also, since each dimension is ensembled with corresponding weights, we could give a very detailed outcome pinning which functions the users are mostly weak in. 

\section{Conclusion \& Future Work}
\
\indent To wrap up, we have developed the framework for a system that offers convenient and accurate PD early-stage diagnosis for the elderly. In the future, we are planning on several things to polish our system:

\begin{enumerate}
    \item Split data collection into two rounds and use first-round data to do some correlation analysis. Combined with direct feedback from users, we are planning to reiterate our APP and model to make it fit into real-world scenarios. 
    \item Add a rehabilitation recommendation system that could arrange rehabilitation plans for the users. The system could also dynamically trace the rehabilitation progress.
    \item Extend Shoupa to more other cerebral diseases, including Alzheimer's and Epilepsy. Since these diseases share similar symptoms, we can use transfer learning to directly transfer data and trained models to other diseases. 
\end{enumerate}

%

\begin{acks}
This work was supported by the 2021-2022 Kunshan Government Research Fund (21KKSGR050). The funders had no role in the study design; data collection, analysis, or interpretation; in the writing of the report; or in the decision to submit the article for publication. This work also obtained IRB Approval from Duke Kunshan University (FWA00021580) and Guangdong General Hospital (KY-Q-2021-208-02).

\end{acks}

\bibliographystyle{unsrt}
\bibliography{sample-base}

\begin{thebibliography}{1}

\bibitem{dorsey2018parkinson}
E~Ray Dorsey and Bastiaan~R Bloem.
\newblock The parkinson pandemic—a call to action.
\newblock {\em JAMA neurology}, 75(1):9--10, 2018.

\bibitem{jankovic2008parkinson}
Joseph Jankovic.
\newblock Parkinson’s disease: clinical features and diagnosis.
\newblock {\em Journal of neurology, neurosurgery \& psychiatry},
  79(4):368--376, 2008.

\bibitem{pinto2004treatments}
Serge Pinto, Canan Ozsancak, Elina Tripoliti, St{\'e}phane Thobois, Patricia
  Limousin-Dowsey, and Pascal Auzou.
\newblock Treatments for dysarthria in parkinson's disease.
\newblock {\em The Lancet Neurology}, 3(9):547--556, 2004.

\bibitem{postuma2015mds}
Ronald~B Postuma, Daniela Berg, Matthew Stern, Werner Poewe, C~Warren Olanow,
  Wolfgang Oertel, Jos{\'e} Obeso, Kenneth Marek, Irene Litvan, Anthony~E Lang,
  et~al.
\newblock Mds clinical diagnostic criteria for parkinson's disease.
\newblock {\em Movement disorders}, 30(12):1591--1601, 2015.

\bibitem{7780459}
Kaiming He, Xiangyu Zhang, Shaoqing Ren, and Jian Sun.
\newblock Deep residual learning for image recognition.
\newblock In {\em 2016 IEEE Conference on Computer Vision and Pattern
  Recognition (CVPR)}, pages 770--778, 2016.

\bibitem{XGboost}
Tianqi Chen and Carlos Guestrin.
\newblock Xgboost: A scalable tree boosting system.
\newblock pages 785--794, 08 2016.

\end{thebibliography}


\end{document}